\documentclass[sigconf]{acmart}
\AtBeginDocument{%
  \providecommand\BibTeX{{%
    \normalfont B\kern-0.5em{\scshape i\kern-0.25em b}\kern-0.8em\TeX}}}
\renewcommand\footnotetextcopyrightpermission[1]{}
\setcopyright{acmcopyright}
\copyrightyear{2022}
\acmYear{2022}
\acmDOI{XXXXXXX.XXXXXXX}
\acmConference[MM '22]{Proceedings of the 30th ACM International Conference on Multimedia}{October 10--14, 2022}{Lisbon, Portugal}
\acmBooktitle{Proceedings of the 30th ACM International Conference on Multimedia (MM '22), October 10--14, 2022, Lisbon, Portugal}
\acmPrice{15.00}
\acmISBN{978-1-4503-XXXX-X/18/06}
\acmPrice{}
\acmISBN{}
\usepackage{multicol}
\usepackage{multirow}
\usepackage{booktabs}
\usepackage{threeparttable}
\usepackage{array}
\usepackage{booktabs}
\usepackage{bbding}
\usepackage{diagbox}
\acmSubmissionID{505}

\begin{document}
\fancyhead{}
\title{NR-DFERNet: Noise-Robust Network for Dynamic Facial Expression Recognition}

\author{Hanting Li, Mingzhe Sui, Zhaoqing Zhu, and Feng Zhao}
\email{{ab828658, sa20, zhaoqingzhu}@mail.ustc.edu.cn,
fzhao956@ustc.edu.cn
}


\begin{abstract}
  Dynamic facial expression recognition (DFER) in the wild is an extremely challenging task, due to a large number of noisy frames in the video sequences. Previous works focus on extracting more discriminative features, but ignore distinguishing the key frames from the noisy frames. To tackle this problem, we propose a noise-robust dynamic facial expression recognition network (NR-DFERNet), which can effectively reduce the interference of noisy frames on the DFER task. Specifically, at the spatial stage, we devise a dynamic-static fusion module (DSF) that introduces dynamic features to static features for learning more discriminative spatial features. To suppress the impact of target irrelevant frames, we introduce a novel dynamic class token (DCT) for the transformer at the temporal stage. Moreover, we design a snippet-based filter (SF) at the decision stage to reduce the effect of too many neutral frames on non-neutral sequence classification. Extensive experimental results demonstrate that our NR-DFERNet outperforms the state-of-the-art methods on both the DFEW and AFEW benchmarks.
\end{abstract}




\maketitle

\section{Introduction}
Facial expressions are the most natural and direct way for humans to express emotions \cite{tian2001}. Understanding human emotional state is a fundamental premise for many computer vision tasks including human-robot interaction (HRI), driver fatigue monitoring and healthcare \cite{li2020,zhang2018}. In recent years, many researchers have developed static facial expression recognition (SFER) methods \cite{li2021mvt,she2021dive,wang2020SCN,xue2021transfer,sui2021ffnet}, which can automatically classify an image into one of seven basic expressions (i.e., neutral, happiness, sadness, surprise, fear, disgust, and anger), and achieved promising results on both lab-controlled (e.g., BU-3DFE \cite{yin2006BU3DFE}) and in-the-wild SFER datasets (e.g., RAF-DB \cite{li2017RAFDB}). Nevertheless, most the natural facial events are dynamic and video can be a better representation of facial expressions. Therefore, DFER has gradually received increasing attention in recent years \cite{liu2014learning,zhao2021former-DFER}.  

According to different data scenarios, DFER datasets can be mainly divided into lab-controlled and in-the-wild. For the former, all the video sequences are collected in a laboratory setting and most sequences show a shift from a neutral facial expression to a peak expression. Over the past decade, many methods are proposed for dealing with lab-controlled DFER task, and achieving promising results \cite{jeong2020ferlab,liu2020dferlab,yu2018dferlab}. Due to such lab-controlled DFER datasets (e.g., CK+ \cite{lucey2010ck+}, MMI \cite{pantic2005mmi}, and Oulu-CASIA \cite{zhao2011Oulu-CASIA}) recording under laboratory conditions, all the facial images are frontal and without any occlusion, leading to a large gap in comparison with the real-world scenario facial images which have variant head poses and different degrees of occlusions. In addition, lab-controlled datasets only contain a very limited number of subjects, which means that multiple videos are often collected from the same subject, this undoubtedly greatly limits the diversity of video sequences. For in-the-wild DFER datasets (e.g., AFEW \cite{dhall2012afew}, Aff-Wild \cite{zafeiriou2017affWILD} and DFEW \cite{jiang2020dfew}), the video sequences are collected from the real-world scenario, which are closer to natural facial events. Besides, in-the-wild datasets are captured from thousands of subjects, which greatly increases the diversity of the data. Moreover, with the collection of the large-scale DFER in-the-wild datasets \cite{jiang2020dfew}, the research focus of DFER has shifted from lab-controlled to challenging in-the-wild conditions. However, due to the existence of variant noisy frames in expression video sequences, the dynamic facial expression recognition (DFER) task which aims to classify a video sequence into a specific expression still remains a significant challenge. 
\begin{figure}[t]
  \centering
  \includegraphics[width=\linewidth]{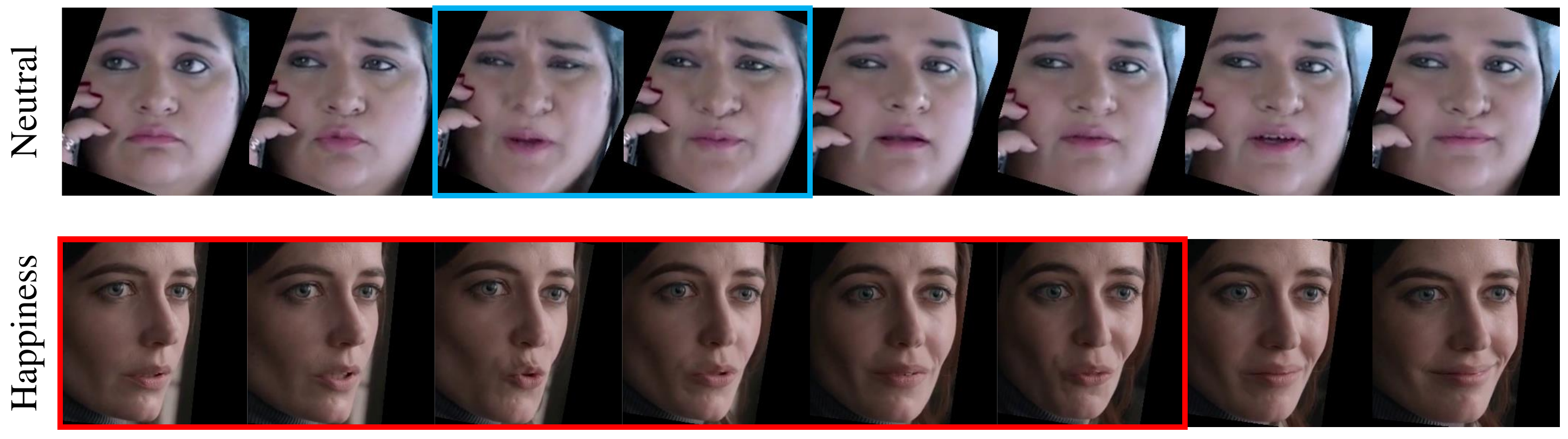}
  \caption{Two kinds of noisy frames in video sequences: (i) $\rm{N_{1}}$: a small amount of target irrelevant frames in video sequence, e.g., frames unrelated to the target expression, frames without face, and heavily occluded frames. (inside the blue box of the top row), and (ii) $\rm{N_{2}}$: excessive neutral frames in non-neutral expression sequence (inside the red box of the bottom row). Both sequences are sampled from DFEW dataset \cite{jiang2020dfew}.}
\vspace{-0.5cm}  
\end{figure}

Before the rise of deep learning, traditional in-the-wild DFER methods are mainly based on hand-crafted features, such as the LBP-TOP \cite{dhall2013LBPTOP}, STLMBP \cite{huang2014STLMBP}, and HOG-TOP \cite{chen2014HOG-TOP}. Besides these hand-crafted descriptors, Liu et al. introduced a spatio-temporal manifold (STM) method to model the video clips \cite{liu2014STM}, and Liu et al. also used different Riemannian kernels measuring the similarity/distance between sequences \cite{liu2014combining}. In recent years, with the development of parallel computing hardware and the collection of large-scale DFER datasets \cite{jiang2020dfew}, deep-learning-based methods have gradually replaced the traditional methods and achieved state-of-the-art performance on in-the-wild DFER datasets. Among variants deep neural networks (DNNs), convolutional neural network (CNNs) are good at modeling spatial features \cite{aminbeidokhti2019cnn,fan2018cnn,meng2019cnn}, recurrent neural networks (RNNs) do well in modeling the temporal relation between frames \cite{lu2018rnn,lee2019rnn,wang2019rnn,ouyang2017rnn,ebrahimi2015rnn}, while 3D CNNs can learn both spatial and temporal features from the sequences \cite{fan20163DCNN,lee20193DCNN,jiang2020dfew}. In order to obtain more emotion-related information, some methods also take multi-modal data (e.g., audio, depth, and thermal recording videos) as the input of the networks \cite{cai2019esmm,lee2020esmm,chen2016esmm,zhang2020esmm,kuhnke2020esmm,mittal2020esmm}.
 Recently, Kumar et al. proposed a novel noisy student training method for DFER \cite{kumar2020noisy}. Jiang et al. collected a large-scale DFER datasets DFEW and introduced a novel expression-clustered spatio-temporal feature learning loss (EC-STFL) for DFER \cite{jiang2020dfew}. Zhao et al. first introduced transformer \cite{vaswani2017transformer} to the DFER task and achieved remarkable results by modeling the spatial and temporal features of the sequences through the proposed convolutional spatial transformer (CS-Former) and temporal transformer (T-Former), respectively \cite{zhao2021former-DFER}. 

Although the above-mentioned methods have gradually improved the performance on in-the-wild DFER datasets, most of them focused on designing or finding a better model to extract more discriminative features, while ignoring distinguishing the key frames and the noisy frames. Unlike lab-controlled sequences that follow a similar pattern (i.e., sequences show a shift from a neutral facial expression to a peak expression), the gap between in-the-wild sequences can be very large, which is also mainly on account of the widespread presence of noisy frames. Among these noisy frames, two kinds of noisy frames can significantly affect the performance of DFER methods. As shown in Figure 1, $\rm{N_{1}}$: a small amount of target irrelevant frames (e.g., frames unrelated to the target expression, frames without face, heavily occluded frames, etc), can interfere with the classification, and $\rm{N_{2}}$: too many neutral frames in non-neutral expression sequences may cause the non-neutral sequences to be misclassified into natural class.


To reduce the interference of these two kinds of noisy frames to DFER, we propose a noise-robust dynamic facial expression recognition network (NR-DFERNet). As shown in Figure 2, our NR-DFERNet can be divided into three stages. In specific, at the spatial stage, we devise a dynamic-static fusion module (DSF) which introduces dynamic features between two adjacent frames as a supplement to the static features, which can make the spatial features more discriminative and robust. Besides, motivated by the success of transformer \cite{vaswani2017transformer} in natural language processing (NLP) tasks and its powerful ability of modeling sequence-based data, we utilize transformer to model the temporal relationship between frames and introduce a novel dynamic class token (DCT) to suppress the impact of target irrelevant frames ($\rm{N_{1}}$) on DFER at the temporal stage. Finally, we propose a snippet-based filter (SF) at the decision stage to reduce the effect of excessive neutral frames ($\rm{N_{2}}$) on non-neutral sequences classification. Extensive experiments indicate that our NR-DFERNet has suppressed the impact of noisy frames on the DFER task and achieved state-of-the-art performance on two popular benchmarks.

In summary, this paper has the following contributions:
\begin{itemize}
	\item We present a noise-robust dynamic facial expression recognition network called NR-DFERNet. The proposed dynamic class token and snippet-based filter suppress the impact of target irrelevant frames ($\rm{N_{1}}$) and too many neutral frames in non-neutral sequences ($\rm{N_{2}}$) on DFER at the temporal and decision stages, respectively. The visualization of the normalized attention weights on each frame show that the proposed NR-DFERNet can pay more attention to the key frames while suppressing the impact of noisy frames.
	
	\item At the spatial stage, we propose a dynamic-static fusion module to obtain more robust and discriminative spatial features from both static features and dynamic features. The visualization of the feature distribution indicate that the proposed NR-DFERNet can learn more discriminative and robust facial features.
	\item  Extensive ablation studies demonstrate the effectiveness of the proposed modules (i.e., DSF, DCT, and SF). Our NR-DFERNet outperforms the baseline model significantly and achieves state-of-the-art results on two popular datasets.
\end{itemize}
\begin{figure*}[t]
  \centering
  \includegraphics[width=\linewidth]{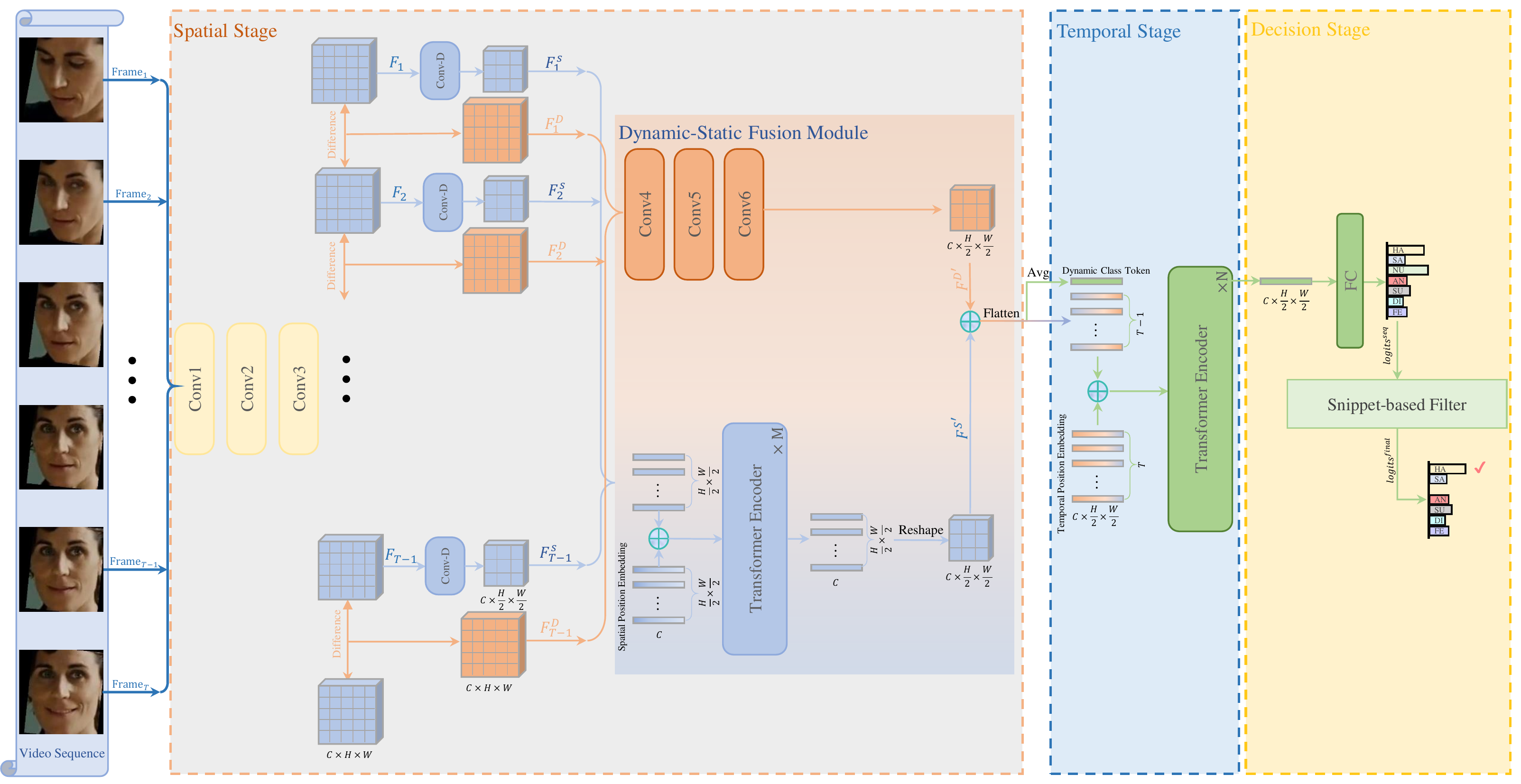}
  \caption{The pipeline of the proposed NR-DFERNet.}\vspace{-0.1cm}
\end{figure*}  
\vspace{-0.5cm}
\section{Related Work}
\subsection{DFER in the Wild}
Before the rise of deep learning, hand-crafted features occupy the mainstream in DFER including LBP-TOP \cite{dhall2013LBPTOP}, STLMBP \cite{huang2014STLMBP}, and HOG-TOP \cite{chen2014HOG-TOP}. However, with the simultaneous development of computing methodologies and computing hardware, deep-learning-based methods gradually replace hand-crafted descriptors in recent years. 

Deep-learning-based methods can be roughly divided into two-stage methods and one-stage methods. In specific, two-stage methods often use CNNs to extract spatial features from frames, then the spatial features are fed into RNNs or transformers to learn the temporal relationships between frames \cite{ebrahimi2015rnn,lee2019rnn,lu2018rnn,ouyang2017rnn,wang2019rnn}. Recently, Liu et al. utilized graph convolutional network (GCN) to learn the frame-based features which focus on a certain expression region \cite{liu2021lstm}. Lee et al. presented multimodal recurrent attention networks (MRAN) to temporally estimate dimensional emotion models by jointly exploiting color, depth, and thermal videos \cite{lee2020esmm}. Zhao et al. first introduced transformer to the DFER task and designed CS-Former and T-Former for extracting spatial and temporal features, respectively \cite{zhao2021former-DFER}. Different from two-stage approaches, one-stage methods often use 3D convolution to learn spatial and temporal features jointly. Kossaifi et al. proposed a tensor factorization framework for efficient separable convolutions of higher order, which allowed training a network on images and using transduction to generalize to videos \cite{kossaifi2020factorized}. Jiang et al. devised a novel EC-STFL loss for DFER in the wild \cite{jiang2020dfew}. The above methods focused on extracting more robust and discriminative spatio-temporal features from video sequences, while ignoring suppressing the noisy frames in the sequence. To address this, we propose NR-DFERNet.

\subsection{Transformer}
Transformers were proposed by Vaswani et al. \cite{vaswani2017transformer} for machine translation. Due to its powerful multi-head self-attention (MHSA) mechanism for modeling long-range dependence, transformers have become the state-of-the-art method in many NLP tasks \cite{otter2020nlpsurvey}. The widespread use of transformers in NLP tasks has attracted great attention in the computer vision (CV) community. Researchers begin to apply transformer-based model in many CV tasks, such as image classification, object detection, and segmentation. Dosovitskiy et al. split an image into $16\times 16$ patches and modeled the relationships between them by a vision transformer (ViT) \cite{dosovitskiy2020vit}. Carion et al. devised a DEtection TRansformer (DETR), which uses object queries to predict the class and bounding box \cite{carion2020DETR}. Wang et al. proposed a pyramid vision transformer (PVT) by incorporating the pyramid structure from CNNs, and achieved good performance on several dense prediction tasks \cite{wang2021pyramid}. As for facial expression recognition, Ma et al. designed visual transformers with feature fusion (VTFF) based on CNN and transformer to learn spatial features of static facial images \cite{ma2021CVT}, and Li et al. proposed a pure transformer-based model called mask vision transformer (MVT), which mask the background of in-the-wild facial images for the SFER task \cite{li2021mvt}. Recently, Zhao et al. devised a dynamic facial expression recognition transformer (Former-DFER) \cite{zhao2021former-DFER} consisting of CS-Former and T-Former for learning spatial and temporal features, respectively.
\vspace{-0.2cm}
\section{Method}
\subsection{Overview}
Our NR-DFERNet is built upon both transformer and CNNs. As shown in Figure 2, NR-DFERNet can be mainly divided into three stages, which are spatial stage, temporal stage, and decision stage. First, the fixed-length facial expression sequence is dynamically sampled from the original video sequence as the input. Then at the spatial stage, several convolutional layers are employed to generate a primary feature map for each frame, and the DSF module containing two branches is proposed to fuse the spatial dynamic features and spatial static features. Next, at the temporal stage, the spatial dynamic-static features are fed into a transformer to model the relationship between the frames and the proposed dynamic class token. Finally, the dynamic class token is mapped to a logits of seven basic expressions by a fully connected (FC) layer, and the final decision is made by the filtered logits through the proposed snippet-based filter.

\subsection{Input Clips}
Our NR-DFERNet takes a clip $X$$\in$$R^{T\times 3\times H_{in}\times W_{in}}$ consisting of $T$ RGB frames as the input, and $X$ is dynamically sampled from the original video. Specifically, we first divide the sequence into $U$ segments equally and then randomly pick $V$ frames from each segment for training. As for the test video sequence, we first split all frames into $U$ segments and then select $V$ frames in the mid of each segment. Hence, the length of the sampled clip is $T=U\times V$ for both the training set and testing set.

After getting a clip $X$ for each video sequence, we feed the clips into three convolutional layers to get $T$ primary feature maps $F_{i}$ with the size of $C\times H\times W$, $i$$\in$$\{1,2,\cdots,T\}$. The dynamic features $F_{i}^{D}$$\in$$R^{C\times H\times W}$ can be represented as,
\begin{equation}\label{1}
F_{i}^{D}=F_{i+1}-F_{i}, \quad\quad\quad\quad\quad i\in \left\{1,2,\cdot\cdot\cdot ,T-1 \right \}.
\end{equation}
and the static features $F_{i}^{S}$$\in$$R^{C\times \frac{H}{2}\times \frac{W}{2}}$ can be formulated as,
\begin{equation}\label{2}
F_{i}^{S}=Conv\mbox{-}D(F_{i}), \quad\quad\quad\quad\quad i\in \left\{1,2,\cdot\cdot\cdot ,T-1 \right \}.
\end{equation}
Where $Conv\mbox{-}D$ is a downsample convolution operation by setting the stride at 2. In particular, since the last frame has no corresponding spatial dynamic feature, we only use the dynamic and static features of the first $T-1$ frames for later feature extraction.

\subsection{Dynamic-Static Fusion Module}
To learn dynamic features and static features separately, our DSF module adopts a two-branch structure, which includes a dynamic branch and a static branch for further learning dynamic features and static features, respectively.

\begin{figure}[t]
  \centering
  \includegraphics[width=\linewidth]{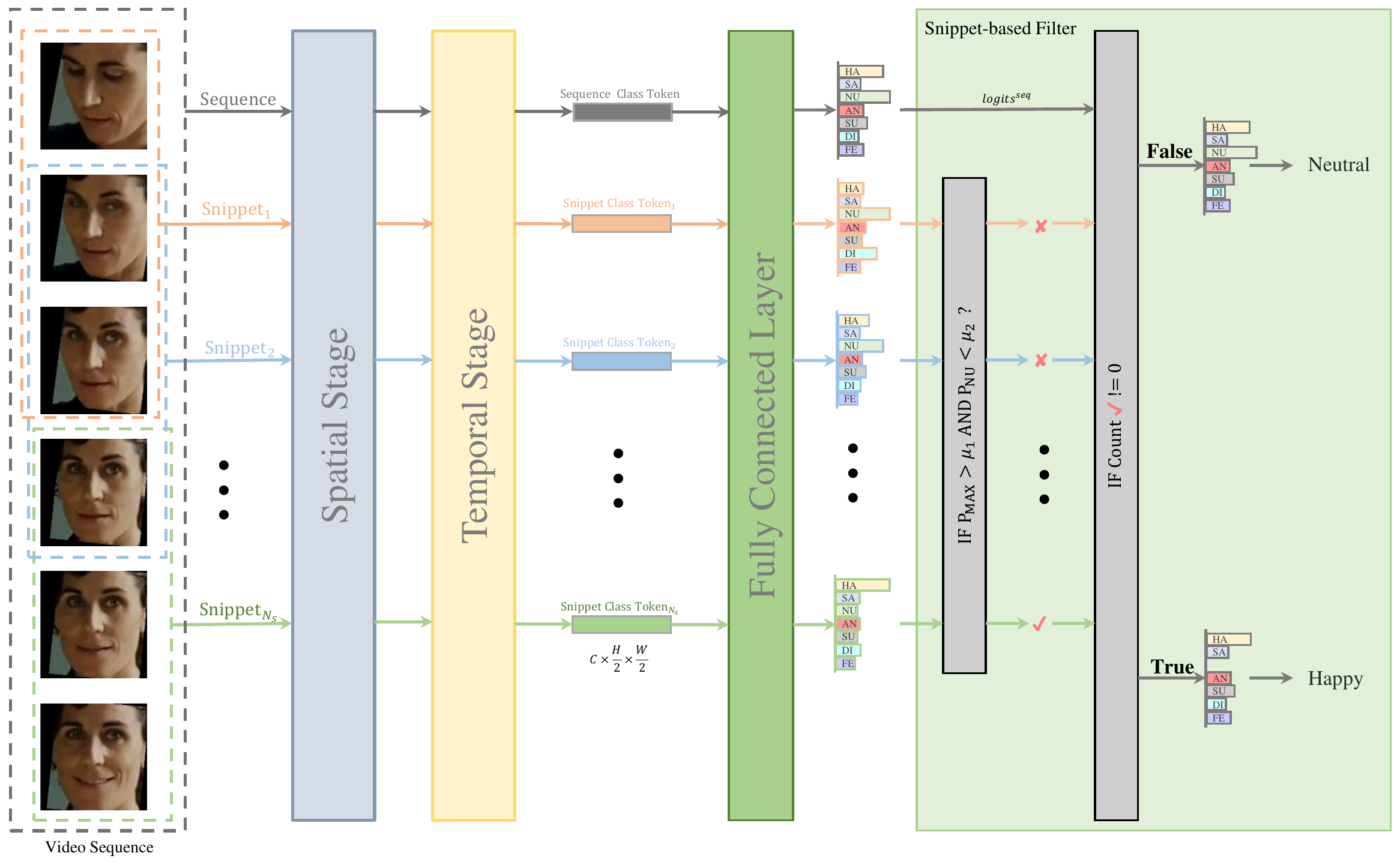}
  \caption{The snippet-based filter.}\vspace{-0.5cm}
\end{figure}
Considering that the differences between adjacent frames are local and sparse, the dynamic features generated from the difference between two feature maps will also be sparse. So we use three convolution layers with $3\times 3$ kernel extract local dynamic features. The dynamic features $F_{i}^{D}$$\in$$R^{C\times H\times W}$ are fed into three convolutional blocks to obtain downsampled dynamic features $F_{i}^{D'}$$\in$$R^{C\times \frac{H}{2}\times \frac{W}{2}}$, $i$$\in$$\{1,2,\cdots,T-1\}$.

For static features, we employ a transformer consisting of $M$ transformer encoders to model the long-range dependence between different parts of the face. In specific, the static feature $F_{i}^{S}$$\in$$R^{C\times \frac{H}{2}\times \frac{W}{2}}$ is flattened into a sequence of $\frac{HW}{4}$ flatten vectors $P_{i}^{S}$ with the size of $C$, $i$$\in$$\{1,2,\cdots,T-1\}$. Then the input embeddings of the transformer $Z^{S}$ can be formulated as,

\begin{equation}\label{3}
Z^{S}=P^{S}+E_{p}^{S}, 
\end{equation}
where $E_{p}^{S}$$\in$$R^{\frac{HW}{4}\times C}$ is a learnable position embeddings for retaining position information \cite{dosovitskiy2020vit}. Then these input embeddings are send to $M$ transformer encoders to learn the relationship between patches. Finally, the output $Z_{i}^{S'}$$\in$$R^{Q\times C}$ ($Q=\frac{H}{2}\times \frac{W}{2}$) is reshaped to $C\times \frac{H}{2}\times \frac{W}{2}$ to get the static features $F_{i}^{S'}$$\in$$R^{C\times \frac{H}{2}\times \frac{W}{2}}$ matching the size of $F_{i}^{D'}$. Finally, the dynamic-static features $F_{i}^{DS}$$\in$$R^{C\times \frac{H}{2}\times \frac{W}{2}}$ can be computed as, 
\begin{equation}\label{4}
F_{i}^{DS}=F_{i}^{D'}+F_{i}^{S'},  \quad\quad\quad\quad\quad i\in \left\{1,2,\cdot\cdot\cdot ,T-1 \right \}.
\end{equation}
By fusing the static features and dynamic features, the dynamic-static features $F_{i}^{DS}$ can provide more discriminative and robust spatial features for the temporal stage.  

\subsection{Dynamic Class Token}
As shown in the top row of Figure 1, there often exists a small amount of target irrelevant frames (e.g., frames unrelated to the target expression, frames without face, frames with heavily occlusion) in video sequences. Learning features from such noisy frames may bring noisy features and affect the classification. Therefore, we introduce a novel dynamic class token which generates a dynamic class token for each sequence to suppress their interference instead of using the same class token for each expression sequence.

Specifically, at the temporal stage, we first reshape dynamic-static features $F_{i}^{DS}$ into the flattened vectors $P_{i}^{DS}$$\in$$R^{\frac{CHW}{4}}$. Then the dynamic class token $T_{d}$$\in$$R^{\frac{CHW}{4}}$ can be defined as,
\begin{equation}\label{4}
T_{d}=\frac{\sum_{i=1}^{T-1}P_{i}^{DS}}{T-1},\quad\quad\quad\quad\quad i\in \left\{1,2,\cdot\cdot\cdot ,T-1 \right \}.
\end{equation}
Based on the assumption that the target irrelevant frames have a large gap with target-related frames, the features of the target irrelevant frames will also have a large gap with that of target-related frames. And in the usual case, when there are only a small number of target irrelevant frames, the DCT $T_{d}$ generated by the mean of the features of each frame should be closer to the features of the target-related frames rather than that of target irrelevant frames. That is to say, in the self-attention mechanism of transformer encoders, DCT will pay more attention to the features of target-related frames, thereby reducing the impact of target irrelevant frames on classification. In contrast, traditional class token in \cite{dosovitskiy2020vit,touvron2021deit}, which is the same for each sequence, cannot adaptively focus on target-related frames in different sequences. 

After adding a learnable position embeddings $E_{p}^{T}$$\in$$R^{T\times \frac{CHW}{4}}$ to the flattened vectors and DCT as we do in the DSF module, the DCT passing through $N$ transformer encoders is mapped to a primary $logits^{seq}$$\in$$R^{7}$, which corresponding to the seven basic expressions.

\subsection{Snippet-based Filter}
The previous DFER works often consider the seven basic expressions (i.e., neutral, happiness, sadness, surprise, fear, disgust, and anger) at the same level while ignoring the special status of neutral expressions. The priority of neutral expressions is lower than other expressions when annotating the expression sequences. For example, when most of the frames in a video sequence are neutral frames and only a small number of frames are non-neutral expressions, the sequence will be finalized as a non-neutral sequence rather than a neutral one. This seems natural to humans, but it is difficult for the network to learn it when given equal-status supervisory information for each expression, which makes the network easily misclassify non-neutral sequences with too many neutral frames into neutral expressions.

To tackle this problem, we design a snippet-based filter to help the proposed NR-DFERNet reduce the effect of excessive neutral frames on non-neutral sequences. As shown in Figure 3, we get $N_{s}$ snippets from the sampled sequence through a sliding window with the width $L$ and stride $S$. For a sampled sequence containing $T$ frames, $N_{s}$ can be formulated as,
\begin{equation}\label{5}
N_{s}=[\frac{T-L}{S}]+1.
\end{equation}
Where ``[]'' represent the round down operation. These snippets will be fed into NR-DFERNet separately to get $N_{s}$ independent logits. These snippet-based logits are then fed into the proposed snippet-based filter together with the primary logits of the whole sequence $logits^{seq}$. The snippet-based filter can be defined as,
\begin{equation}\label{6}\small
\begin{aligned}
logits^{final}=\begin{cases}
 &logits^{seq} \textbf{ if } \forall i \quad s.t. \quad P_{MAX}^{i}<\mu _{1} \lor P_{NU}^{i}>\mu _{2},\\
 & logits^{seq'} \textbf{ if } \exists i \quad s.t. \quad P_{MAX}^{i}>\mu _{1} \land P_{NU}^{i}<\mu _{2}.
\end{cases}
\end{aligned}
\end{equation}

Where $\land$ and $\lor$ represents ``and'' and ``or'', respectively. $logits^{final}$ stands for the filtered logits used for making the final decision, $P_{MAX}^{i}$ and $P_{NU}^{i}$, $i$$\in$$\{1,2,\cdot\cdot\cdot, N_{s}\}$, stand for the maximum predicted probability and the predicted probability of neutral of the i-th snippet, respectively. $\mu_{1}$ and $\mu_{2}$ are two thresholds for logical judgment, while $logits^{seq}$ denotes the original logits of the sequence and $logits^{seq'}$ represents $logits^{seq}$ with the predicted probability of neutral class setting to zero as described in Figure 3. In other words, when there exists a snippet in a sequence that points to a clear non-neutral expressions, even if other snippets are all neutral expressions, the sequence is preferentially classified as a non-neutral expression at the decision stage.

\section{Experiments}
We carry out extensive experiments on two popular in-the-wild DFER datasets (i.e., DFEW \cite{jiang2020dfew} and AFEW \cite{dhall2012afew}). In this section, we first introduce the datasets and implementation details. Then we explore the effectiveness of each component of NR-DFERNet on DFEW dataset. Subsequently, we compare the proposed method with several state-of-the-art approaches and give some visualizations to demonstrate the effectiveness of our NR-DFERNet.
\subsection{Datasets}
\textbf{DFEW} \cite{jiang2020dfew} consists of over 16,000 video clips from thousands of movies, which is the current largest in-the-wild benchmark for DFER. These video clips contain various challenging interferences in practical scenarios such as extreme illumination, occlusions, and capricious pose changes. In addition, each video clip is individually annotated by ten independent individuals under professional guidance and assigned to one of seven basic expressions, (i.e., happiness, sadness, neutral, anger, surprise, disgust, and fear). Consistent with the previous works \cite{zhao2021former-DFER,jiang2020dfew}, we only conduct experiments on 12059 video clips, which can be clearly assigned to a specific single-labeled emotion category. All the samples have been split into five same-size parts (fd1\textasciitilde fd5) without overlap. So we choose 5-fold cross-validation, which takes one part of the samples for testing and the remaining for training, as the evaluation protocol. 

\textbf{AFEW} \cite{dhall2012afew} dataset served as an evaluation platform for the annual EmotiW from 2013 to 2019. AFEW contains 1809 video clips collected from different movies and TV serials, and all the samples have been split into three subsets: training (773 video clips), validation (383 video clips), and testing (653 video clips) set. Consistent with the DFEW, each video clip in AFEW is assigned to one of seven basic expressions. Since the testing set is not publicly available, we train our model on the training set and report results on the validation set as the previous methods did. 

\subsection{Implementation Details}
\textbf{Preprocessing:} For AFEW dataset, we detect the face regions of each frame via the Retinaface \cite{deng2020retinaface}, and then the face region is cropped and aligned according to the bounding box and landmarks. Specially, due to the low-light issues that existed in AFEW, a pre-trained deep learning model Enlighten-GAN \cite{jiang2021enlightengan}, is used to enhance the light, which is consistent with \cite{zhao2021former-DFER}. For DFEW dataset, the face regions of the video frames are publicly available, so we directly use the processed data for experiments. 
\begin{table}
\begin{center}

\label{table:headings}\caption{Evaluation of each component in NR-DFERNet (i.e., dynamic-static fusion module, dynamic class token and snippet-based filter). The best results are in bold.}\vspace{-0.3cm}
\begin{tabular}{c|c|c|c|cc}
\toprule
\multirow{2}{*}{Setting}&\multicolumn{2}{c|}{Feature Level}&\multicolumn{1}{c|}{Decision Level}&\multicolumn{2}{c}{ Metrics (\%)}\cr
    \cmidrule(lr){2-6}&DSF & DCT & SF & UAR & WAR\cr
\midrule
a (Baseline)&              &                &             &50.22 & 63.52\cr
b&              &                & \CheckmarkBold  &50.79 &63.85  \cr
c&              & \CheckmarkBold &                &51.21 &64.62 \cr
d&              &\CheckmarkBold  &\CheckmarkBold &51.32 &64.90  \cr
e&\CheckmarkBold&                &               &52.35 & 65.36 \cr
f&\CheckmarkBold&                &\CheckmarkBold  &52.76 &66.02 \cr
g&\CheckmarkBold& \CheckmarkBold &              &53.50 &67.31  \cr
\midrule
h&\CheckmarkBold&\CheckmarkBold  &\CheckmarkBold &\textbf{54.21}&\textbf{68.19}\cr

\bottomrule

\end{tabular}
\vspace{-0.5cm}
\end{center}
\end{table}

\textbf{Training Setting:} In our experiments, facial images are all resized to the size of 112 $\times$ 112. The random cropping, horizontal flipping, and color jittering are employed to avoid over-fitting. We use SGD \cite{robbins1951sgd} to optimize NR-DFERNet with a batch size of 32. For DFEW dataset, the learning rate is initialized to 0.001, decreased at an exponential rate in 70 epochs for cross-entropy loss function. For AFEW dataset, the learning rate is initialized to 0.0003, also decreased at an exponential rate in 30 epochs for cross-entropy loss function. Due to the tiny quantity of the data samples in AFEW, previous work pre-trained their models on different datasets (both static and dynamic FER datasets). Consistent with \cite{zhao2021former-DFER}, we first pre-train our NR-DFERNet and other models on DFEW (fd1) and then fine-tune on AFEW. For both datasets, in our method, $U$= 8, $V$= 2. Hence, the length of the dynamically sampled sequence is 16. And the number of all the self-attention heads is set at 4. It is worth noting that the results are not sensitive to $\mu_{1}$ and $\mu_{2}$ of the snippet-based filter, so we present our ablation experiments on them in the supplementary material. By default, $\mu_{1}$ and $\mu_{2}$ are set to 0.7 and 0.05, respectively. All the experiments are conducted on a single NVIDIA RTX 3070 card with PyTorch toolbox \cite{paszke2019pytorch}.

\textbf{Validation Metrics:} Consistent with the EC-STFL \cite{jiang2020dfew} and Former-DFER \cite{zhao2021former-DFER}, we choose the unweighted average recall (UAR, i.e., the accuracy per class divided by the number of classes without considering the number of instances per class) and weighted average recall (WAR, i.e., accuracy) as the metrics.

\subsection{Ablation Studies}
As shown in Figure 2, NR-DFERNet mainly consists of three key components, i.e., dynamic-static fusion module, dynamic class token, and snippet-based filter at three different stages. To show the effectiveness of our NR-DFERNet, we conduct ablation studies to evaluate the influence of the key hyper-parameters and components on the DFEW \cite{jiang2020dfew} dataset with 5-fold cross-validation.
\begin{table}
\begin{center}

\label{table:headings}\caption{Evaluation of the stride \textit{S} and window width \textit{L} of snippet-based filter. The best results are in bold.
}
\vspace{-0.3cm}
\begin{tabular}{c|c|cc}
\toprule
\multicolumn{2}{c|}{Setting}&\multicolumn{2}{c}{ Metrics (\%)}\cr
    \cmidrule(lr){1-4}Stride&Window Width
    & UAR & WAR\cr
\midrule
1 &3 &54.16&68.12 \cr
2 &3 &\textbf{54.21}&\textbf{68.19}  \cr
3 &3 &54.08&68.06\cr
1 &6 &54.02&67.94\cr
2 &6 &53.96&67.89 \cr
3 &6 &54.04&68.02\cr
\bottomrule
\end{tabular}
\vspace{-0.5cm}
\end{center}
\end{table}

\begin{table}
\begin{center}

\label{table:headings}\caption{Evaluation of the number of the transformer encoder layers. M and N stands for volumes of the transformer encoders at spatial stage and temporal stage, respectively. The best results are in bold.
}
\vspace{-0.3cm}
\begin{tabular}{c|c|cc|c}
\toprule
\multicolumn{2}{c|}{Setting}&\multicolumn{2}{c|}{ Metrics (\%)}&\multirow{2}{*}{FLOPs}\cr
    \cmidrule(lr){1-4}M&N
    & UAR & WAR&(G)\cr
\midrule
1 &4 &54.02 &68.04&\textbf{6.20} \cr
2 &4 &\textbf{54.21}&\textbf{68.19}&6.33  \cr
3 &4 &53.89 &67.86&6.46\cr
1 &6 &53.30&67.31&7.31\cr
2 &6 &53.72&67.52&7.44 \cr
3 &6 &53.05&67.12&7.57\cr
\bottomrule
\end{tabular}
\vspace{-0.5cm}
\end{center}
\end{table}
\textbf{Evaluation of Each Component:} We first study the effectiveness of each component in our NR-DFERNet in Table 1. In specific, for building the baseline model, we only use the static features $F_{i}^{S'}$ in Eq. (4) to represent the spatial features for each frame, employ the original class token used in \cite{dosovitskiy2020vit} to aggregate features of each frame, and directly make the final decision based on the unfiltered logits $logits^{seq}$. From setting (a, e), setting (b, f) and setting (d, h), we can clearly see that the dynamic-static fusion module can improve the performance of both WAR and UAR. When we employ DCT to suppress the impact of target irrelevant frames, we can see that DCT improve the performance by 1.15\%/1.95\% of UAR/WAR from setting (e, g). In general, our NR-DFERNet setting (h) exceeds the baseline setting (a) by 3.99\%/4.67\% of UAR/WAR, which fully indicates the importance and effectiveness of the proposed modules.


\textbf{Outlook for Snippet-based Filter:} From Table 1, we can see that compared to the DSF module and DCT, the snippet-based filter seems to have a limited improvement in performance. Since the proposed snippet-based filter is at the decision stage, we believe that the quality of snippet-based logits can directly affect the performance of SF. To confirm this, we present the performance curve (WAR) of the model on the test set with and without SF in Figure 4. In addition, we also present the performance gain brought by SF for a more intuitive observation. It is obvious that the gain increase with the improvement of the overall performance in the trend, which is predictable because when we have more accurate analyses for the snippets and the whole sequences, the SF that operates based on these analyses can also function better accordingly. This also proves that it is beneficial for models to consider neutral frames in a special position for DFER, and we believe that the gain brought by SF will expand as the model performance is further improved in the future.
\begin{figure}[t]
  \centering
  \includegraphics[width=7cm]{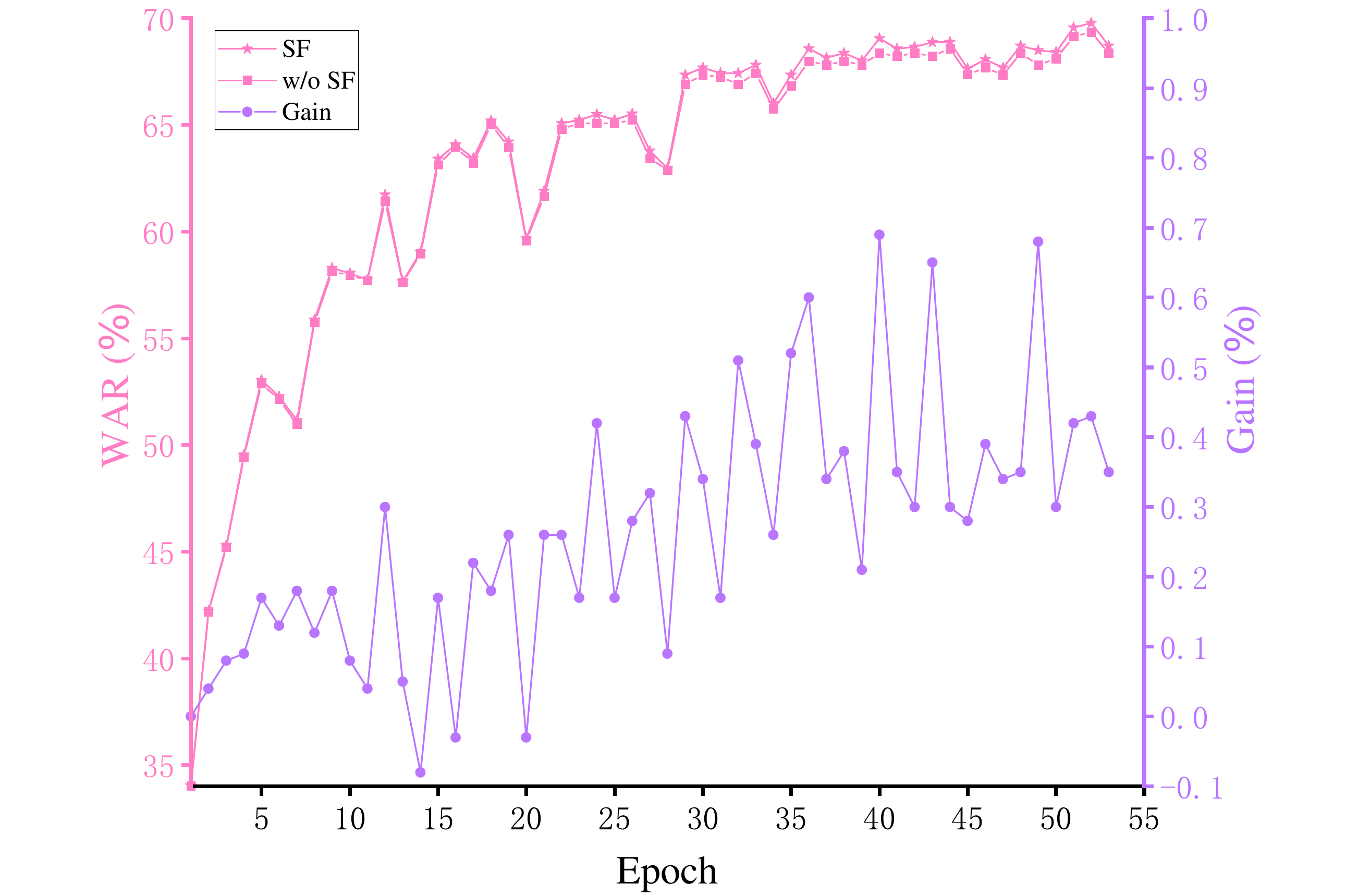}
  \caption{WAR and gain curve of the snippet-based filter on fd5 of DFEW (The window width $L$ and the stride $S$ are set at 3 and 2, respectively). }
  \vspace{-0.2cm}
\end{figure}

\begin{table}
\begin{center}

\label{table:headings}\caption{Comparison with state-of-the-art methods on AFEW. DS denotes dynamic sampling. The bold denotes the best.
}\vspace{-0.3cm}
\begin{tabular}{c|c|cc|c}
\toprule
\multirow{2}{*}{Method}&\multirow{2}{*}{Sample}&\multicolumn{2}{c|}{ Metrics (\%)}&\multirow{2}{*}{FLOPs}\cr
    \cmidrule(lr){3-4}&Strategy
    & UAR & WAR&(G)\cr
\midrule
EmotiW-2019 Baseline \cite{dhall2019emotiw} &n/a &n/a &38.81&n/a \cr
C3D\cite{tran2015C3D} &DS &43.75 &46.72&38.57 \cr
I3D-RGB \cite{carreira2017I3D-RGB}&DS &41.86 &45.41&6.99 \cr
R(2+1)D \cite{tran2018closer}&DS &42.89 &46.19&42.36 \cr
3D ResNet18 \cite{hara20183dresnet}&DS &42.14 &45.67&8.32 \cr
ResNet18+LSTM \cite{hochreiter1997lstm,he2016resnet}&DS &43.96 &48.82&7.78 \cr
ResNet18+GRU \cite{he2016resnet,chung2014gru}&DS &45.12 &49.34&7.78 \cr
EC-STFL \cite{jiang2020dfew}&DS&47.33 &50.66&8.32\cr
Former-DFER \cite{zhao2021former-DFER}&DS &47.42 &50.92&9.11 \cr
\midrule
NR-DFERNet (Ours)&DS&\textbf{48.37}&\textbf{53.54}&\textbf{6.33}\cr
\bottomrule
\end{tabular}
\vspace{-0.5cm}
\end{center}
\end{table}

\begin{table*}
\begin{center}

\label{table:headings}\caption{Comparison with state-of-the-art methods on DFEW. TI denotes time interpolation \cite{zhou2011ti,zhou2013ti}. DS denotes dynamic sampling. $^\ast$Oversampling is used since the DFEW dataset is imbalanced. The bold denotes the best.
}\vspace{-0.3cm}
\begin{tabular}{c|c|ccccccc|cc|c}
\toprule
\multirow{2}{*}{Method}&\multirow{2}{*}{Sample}&
    \multicolumn{7}{c|}{Accuracy of Each Emotion (\%) }&\multicolumn{2}{c|}{ Metrics (\%)}&\multirow{2}{*}{FLOPs}\cr
    \cmidrule(lr){3-11}&
    Strategy&Happiness & Sadness & Neutral & Anger & Surprise & Disgust & Fear & UAR & WAR&(G)\cr
\midrule
C3D \cite{tran2015C3D}&TI &75.17 &39.49 &55.11 &62.49 &45.00 &1.38 &20.51 &42.74 &53.54& 38.57\cr
P3D \cite{qiu2017p3d} &TI &74.85 &43.40 &54.18 &60.42 &50.99 &0.69 &23.28 &43.97 &54.47& n/a\cr
R(2+1)D18 \cite{tran2018closer} &TI &79.67 &39.07 &57.66 &50.39 &48.26 &3.45 &21.06 &42.79 &53.22&42.36 \cr
3D Resnet18 \cite{hara20183dresnet}&TI &73.13 &48.26 &50.51 &64.75 &50.10 &0.00 &26.39 &44.73 &54.98&8.32 \cr
I3D-RGB \cite{carreira2017I3D-RGB}&TI &78.61 &44.19 &56.69 &55.87 &45.88 &2.07 &20.51 &43.40 &54.27&6.99 \cr
VGG11+LSTM \cite{simonyan2014vgg,hochreiter1997lstm}&TI &76.89 &37.65 &58.04 &60.70 &43.70 &0.00 &19.73 &42.39 &53.70&31.65 \cr
ResNet18+LSTM \cite{hochreiter1997lstm,he2016resnet}&TI &78.00 &40.65 &53.77 &56.83 &45.00 &\textbf{4.14} &21.62 &42.86 &53.08&7.78 \cr
3D R.18+Center Loss \cite{hara20183dresnet,wen2016discriminative}&TI &78.49 &44.30 &54.89 &58.40 &52.35 &0.69 &25.28 &44.91 &55.48&8.32 \cr
EC-STFL \cite{jiang2020dfew}&TI &79.18 &49.05 &57.85 &60.98 &46.15 &2.76 &21.51 &45.35 &56.51&8.32 \cr
3D Resnet18 \cite{hara20183dresnet}&DS &76.32 &50.21 &64.18 &62.85 &47.52 &0.00 &24.56 &46.52 &58.27&8.32 \cr
ResNet18+LSTM \cite{hochreiter1997lstm,he2016resnet}&DS &83.56 &61.56 &68.27 &65.29 &51.26 &0.00 &29.34 &51.32 &63.85&7.78 \cr
Resnet18+GRU \cite{he2016resnet,chung2014gru}&DS &82.87 &63.83 &65.06 &68.51 &52.00 &0.86 &30.14 &51.68 &64.02&7.78 \cr
Former-DFER \cite{zhao2021former-DFER}&DS &84.05 &62.57 &67.52 &70.03 &56.43 &3.45 &31.78 &53.69 &65.70&9.11 \cr
\midrule
NR-DFERNet (Ours)&DS&\textbf{88.47}&64.84&70.03&\textbf{75.09}&\textbf{61.60}&0.00&19.43&54.21&\textbf{68.19}&\textbf{6.33}\cr
NR-DFERNet$^\ast$ (Ours)&DS&86.42&\textbf{65.10}&\textbf{70.40}&72.88&50.10&0.00&\textbf{45.44}&\textbf{55.77}&68.01&\textbf{6.33}\cr
\bottomrule
\end{tabular}
\end{center}
\end{table*}

\begin{figure*}[ht]
  \centering
  \includegraphics[width=\linewidth]{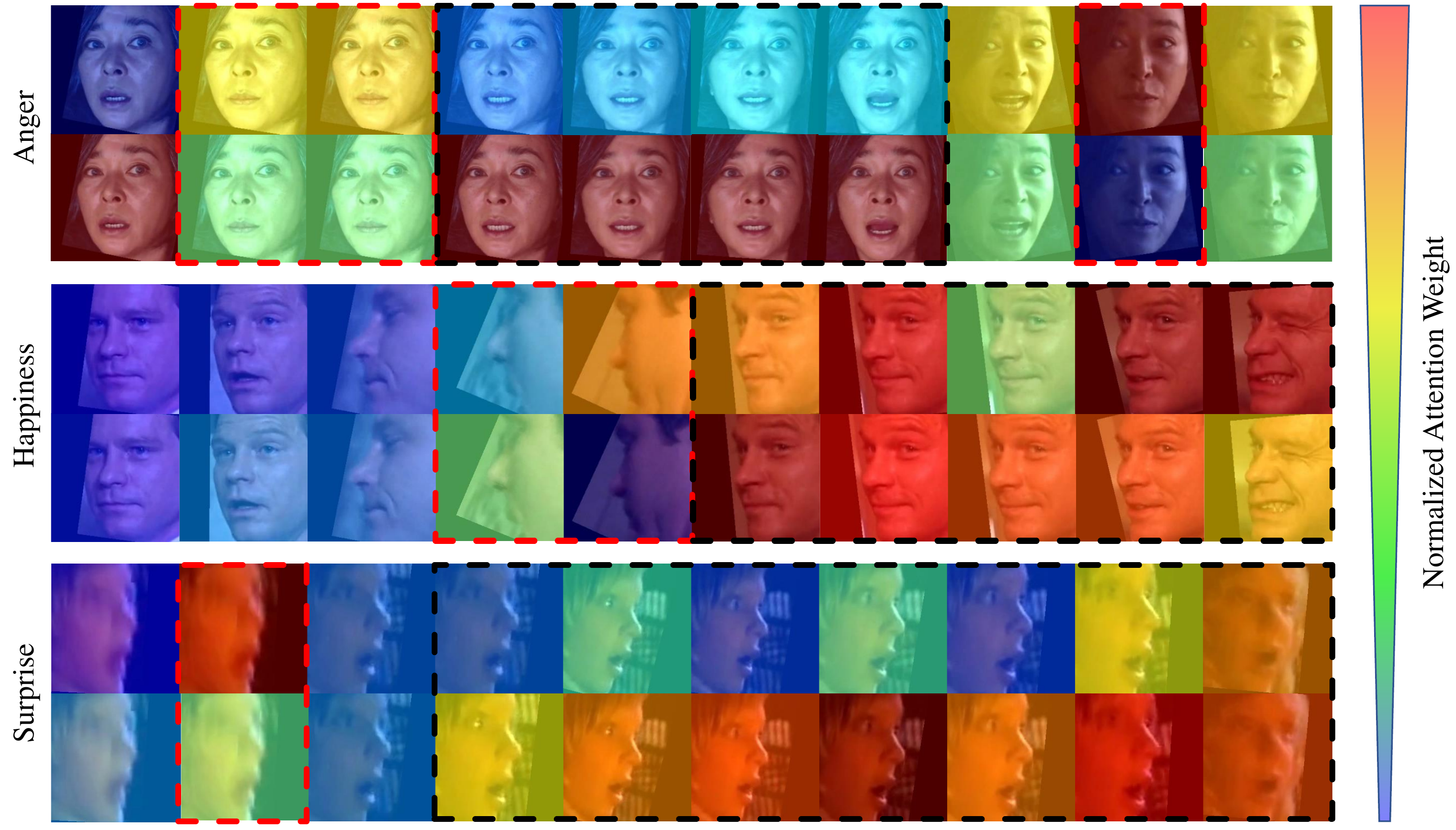}
  \caption{Visualization of the normalized attention weights of the transformer at the temporal stage. For each sequence, the images in the first row are the normalized attention weights generated by NR-DFERNet without the dynamic class token, and the images in the second row are the normalized attention weights generated by NR-DFERNet with dynamic class token. The frames inside the red boxes stand for the target irrelevant frames (i.e., $\rm{N_{1}}$ in Figure 1), while the frames inside the black box represent the target-related frames.}
  
\end{figure*}
\begin{figure*}[h]
  \centering
  \includegraphics[width=\linewidth]{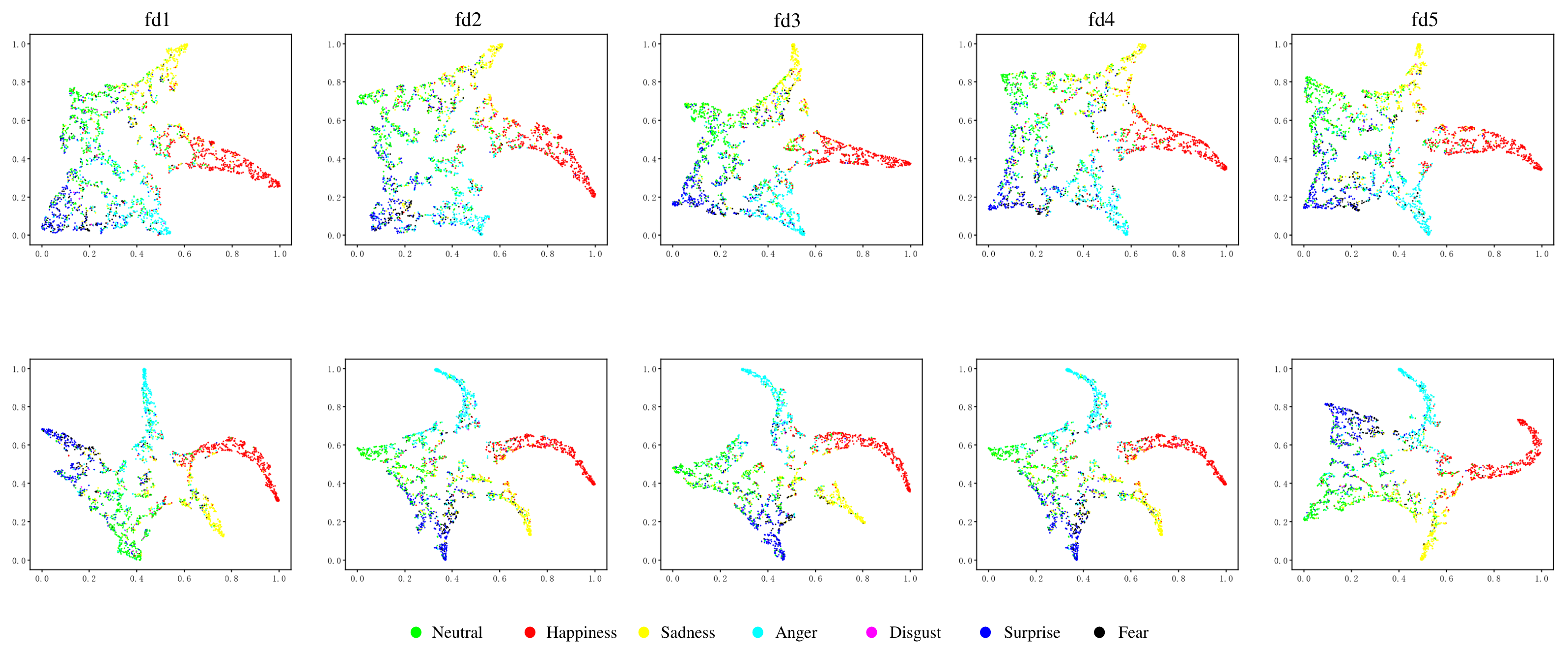}
  \caption{Feature distribution learned by the baseline (top) and our NR-DFERNet (bottom) on fd1\textasciitilde fd5 of DFEW benchmark.}
  
\end{figure*}

\textbf{Evaluation of Hyper-parameters:} We conduct ablation studies on several key hyper-parameters of NR-DFERNet to explore their influence on the performance. For the snippet-based filter, Table 2 shows the performance of the model under different sliding window settings. The results show that smaller window width which corresponds to a shorter snippet can bring an bigger increase in both UAR and WAR. This is reasonable because shorter snippets correspond to more fine-grained local expression information, which can help snippet-based filter to find snippets with clear non-neutral expression snippets in a large number of neutral frames. Since the sliding window has translation invariance, the stride only has little effect on the final performance. The stride and window width are set to 2 and 3 by default.

We also study the effect of the different number of layers for the transformer encoders at the spatial stage and temporal stage, respectively. As shown in Table 3, a transformer encoder that is too shallow has limited fitting ability and cannot fit the data well. A too deep transformer is prone to overfitting when the training data are limited. So the default number of layers for transformers at spatial and temporal stages are set to 2 and 4, respectively.

\vspace{-0.4cm}
\subsection{Comparison with State-of-the-Arts}
In this section, we compare our best results with several state-of-the-art methods on the AFEW and DFEW benchmarks.

\textbf{Results on AFEW:} For a fair comparison, all the models in Table 4 are first pre-trained on DFEW (fd1) and then fine-tuned on AFEW as \cite{zhao2021former-DFER} did. As shown in Table 4, under the same sampling strategy, our NR-DFERNet exceeds Former-DFER by 0.95\%/2.62\% of UAR/WAR with lower FLOPs, which demonstrate the effectiveness of the proposed method.

\textbf{Results on DFEW:} We conduct 5-fold cross-validation as previous works did on DFEW dataset. All the experiments including ablation studies of our NR-DFERNet are carried out under a dynamic sampling strategy, so we also experiment with some previous methods using dynamic sampling as a comparison. From Table 5, we can see that our NR-DFERNet outperforms the best results of previous methods on every evaluation metric with the lowest FLOPs. Concretely, our method outperforms the previous state-of-the-art method Former-DFER \cite{zhao2021former-DFER} by 0.52\%/2.49\% of UAR/WAR. From Table 5, we can see that the poor performance on “disgust” and “fear”, we think that the severe class imbalance of DFEW results in poor performance. Specifically, for the DFEW dataset, the proportions of ``disgust'' and ``fear'' sequences are 1.22\% and 8.14\%, respectively. So we mitigate the impact of class imbalance by using oversampling strategy (i.e., reduce the sampling of classes with a large proportion and increase the sampling of classes with a small proportion), which significantly improves the performance of UAR with a small loss of WAR. In specific, our NR-DFERNet exceeds Former-DFER by 2.08\%/2.31\% of UAR/WAR with oversampling strategy.

\vspace{-0.3cm}

\subsection{Visualization}
To prove that the proposed dynamic class token can suppress the impact of the target irrelevant frames, we multiply the attention map in each layer of the temporal stage transformer and normalize it to obtain the normalized attention weight for each frame of the sampled sequence. We visualize the normalized attention weights of the multi-head self-attention layers in Figure 5, which reflects how much attention the class token paid to each frame. In specific, for the target irrelevant frames in the red box, the proposed DCT can help the attention mechanism pay less attention to these noisy frames. While for the target-related frames in the black box, our DCT can give more stable and continuous attention to these key frames, which fully demonstrates the effectiveness of the DCT.

Moreover, we also utilize t-SNE \cite{van2008tsne} to analyze the feature distribution learn by the baseline and our NR-DFERNet on five folds of DFEW (fd1\textasciitilde fd5). As shown in Figure 6, it is obvious that the features of each class learned by NR-DFERNet are more closely distributed, and the boundaries between different classes are more pronounced. This shows that our NR-DFERNet can extract more discriminative features from the sequence.

We also present the confusion matrix in the supplementary material to demonstrate that the proposed SF do reduce the number of non-neutral samples misclassified as neutral expression.

\vspace{-0.5cm}
\section{Conclusion}
In this paper, we develop a noise-robust dynamic facial expression recognition network for suppressing the impact of two kinds of noisy frames (i.e., $\rm{N_{1}}$ and $\rm{N_{2}}$ in Figure 1), at the temporal and the decision stages, respectively. Specifically, we devise a dynamic-static fusion module to get more discriminative and robust spatial features by learning from both dynamic features and static features. Besides, we introduce a novel dynamic class token to suppress the impact of target irrelevant frames at the temporal stage. Finally, at the decision stage, we design a snippet-based filter considering the special status of neutral expressions in the DFER task, which further improves the performance of the model. Extensive experiments demonstrate that our NR-DFERNet outperforms other state-of-the-art methods on two popular benchmarks.
\bibliographystyle{ACM-Reference-Format}
\bibliography{sample-sigconf}


\end{document}